\documentclass[11pt,letterpaper]{article}
\usepackage{emnlp2017}
\usepackage{times}
\usepackage{latexsym}

\usepackage{float}
\usepackage{bm}
\usepackage[pdftex]{graphicx} 
\usepackage{amsmath}
\usepackage{amsfonts}
\usepackage{multirow}
\usepackage{booktabs}
\usepackage{arydshln}
\usepackage{flushend}
\usepackage{pgfplots}
\usepackage{caption}

\emnlpfinalcopy

\def\emnlppaperid{1249}

\newcommand{\squishlist}{
 \begin{list}{$\bullet$}
  { \setlength{\itemsep}{0pt}
     \setlength{\parsep}{3pt}
     \setlength{\topsep}{3pt}
     \setlength{\partopsep}{0pt}
     \setlength{\leftmargin}{1.5em}
     \setlength{\labelwidth}{1em}
     \setlength{\labelsep}{0.5em} } }

\newcommand{\squishlisttwo}{
 \begin{list}{$\bullet$}
  { \setlength{\itemsep}{0pt}
    \setlength{\parsep}{0pt}
    \setlength{ opsep}{0pt}
    \setlength{\partopsep}{0pt}
    \setlength{\leftmargin}{2em}
    \setlength{\labelwidth}{1.5em}
    \setlength{\labelsep}{0.5em} } }

\newcommand{\squishend}{
  \end{list}  }

\pgfplotstableread[col sep=comma]{
0.2,0.8538,0.8618,0.8642,0.8625,0.8716,0.8722,0.8679,0.8757 
0.4,0.8579,0.8622,0.8655,0.8715,0.8732,0.8769,0.866,0.8783 
0.6,0.8637,0.8635,0.8672,0.8762,0.8766,0.8798,0.8681,0.8843 
0.8,0.8643,0.8724,0.8781,0.8801,0.8832,0.8822,0.8786,0.8946 
1,0.8651,0.8752,0.8791,0.8816,0.8863,0.8834,0.8795,0.8952 
}\imdbyelp
\pgfplotstableread[col sep=comma]{
0.2,0.7636,0.7721,0.7752,0.7721,0.7812,0.7875,0.7903,0.7861
0.4,0.7671,0.7754,0.7778,0.7772,0.7985,0.7924,0.7911,0.8059
0.6,0.7709,0.7789,0.7791,0.7785,0.8052,0.7946,0.7944,0.815
0.8,0.7739,0.7812,0.7842,0.7882,0.8122,0.7959,0.8062,0.8307
1,0.7742,0.7832,0.7868,0.7912,0.8177,0.7961,0.8169,0.8362
}\yelpimdb
\title{Learning Word Embeddings with Text from Different Domains}
\title{Exploiting Multiple Domains for Learning Word Embeddings}
\title{Better Cross-Domain Word Representation Learning}
\title{A Simple and Effective Algorithm for Learning Cross-Domain \\
Word Embeddings}

\title{A Simple Regularization-based \\Algorithm for Learning Cross-Domain 
Word Embeddings}

\author{Wei Yang \\  University of Waterloo \\  \texttt{w85yang@uwaterloo.ca}
        \And  Wei Lu  \\ Singapore University \\ of Technology and Design \\  \texttt{luwei@sutd.edu.sg}\And
        Vincent W. Zheng  \\  Advanced Digital Sciences Center \\  \texttt{vincent.zheng}\\\texttt{@adsc.com.sg} }
        
\date{}

\begin{document}

\maketitle

\begin{abstract}
Learning word embeddings has received a significant amount of attention recently.
Often, word embeddings are learned in an unsupervised manner from a large collection of text. 
The genre of the text typically plays an important role in the effectiveness of the resulting embeddings.
How to effectively train word embedding models using data from different domains remains a problem that is underexplored.
In this paper, we present a simple yet effective method for learning word embeddings based on text from different domains.
We demonstrate the effectiveness of our approach through extensive experiments on various down-stream NLP tasks.
\end{abstract}

\section{Introduction}

Recently, the learning of distributed representations for natural language words (or word embeddings) has received a significant amount of attention \cite{mnih2007three,turian2010word,mikolov2013efficient,mikolov2013distributed,mikolov2013linguistic,pennington2014glove}.
Such representations were shown to be able to capture syntactic and semantic level information associated with words \cite{mikolov2013efficient}.
Word embeddings were shown effective in tasks such as named entity recognition \cite{sienvcnik2015adapting}, sentiment analysis \cite{lihao2017sentiment} and syntactic parsing \cite{durrett2015neural}. One common assumption made by most of the  embedding methods is that, the text corpus is from one single domain; e.g., articles from bioinformatics. 
However, in practice, there are often text corpora from multiple domains; e.g., we may have text collections from broadcast news or Web blogs, whose words are not necessarily limited to bioinformatics. 
Can these corpora from different domains help learn better word embeddings, so as to improve the downstream NLP applications in a target domain like bioinformatics? Our answer is yes, because despite the domain differences, these additional domains do introduce more text data converying useful information (i.e., more words, more word co-occurrences), which can be helpful for consolidating the word embeddings in the target bioinformatics domain.



In this paper, we propose a simple and easy-to-implement approach for learning cross-domain word embeddings. Our model can be seen as a regularized \emph{skip-gram} model \cite{mikolov2013efficient,mikolov2013distributed}, where the source domain information is selectively incorporated for learning the target domain word embeddings in a principled manner. 


\section{Related Work}

Learning a continuous representation for words has been studied for quite a while \cite{Hinton:1986:DR:104279.104287}. Many earlier word embedding methods employed the computationally expensive neural network architectures \cite{collobert2008unified,mikolov2013linguistic}.
Recently, an efficient method for learning word representations, namely the {skip-gram} model {\cite{mikolov2013efficient,mikolov2013distributed}} was proposed and implemented in the widely used word2vec toolkit. It tries to use the current word to predict the surrounding  context words, where the prediction is defined over the embeddings of these words. As a result, it learns the word embeddings by maximizing the likelihood of predictions. 

Domain adaptation is an important research topic \cite{pan2013transfer}, and it has been considered in many NLP tasks. For example, domain adaptation is studied for sentiment classification \cite{glorot2011domain} and parsing \cite{mcclosky2010automatic}, just to name a few. 
However, there is very little work on domain adaptation for word embedding learning. One major reason preventing people from using text corpora from different domains for word embedding learning is the lack of guidance on which kind of information is worth learning from the source domain(s) for the target domain. In order to address this problem, some pioneering work has looked into this problem.
 For example, \newcite{bollegala2015unsupervised} considered those frequent words in the source domain and the target domain as the ``pivots''. Then it tried to use the pivots to predict the surrounding ``non-pivots'', meanwhile ensuring the pivots to have the same embedding across two domains. 
 Embeddings learned from such an approach were shown to be able to improve the performance on a cross-domain sentiment classification task.
However, this model fails to learn embeddings for many words which are neither pivots nor non-pivots, which could be crucial for some downstream tasks such as named entity recognition.




\section{Our Approach}


Let us first state the objective of the skip-gram model \cite{mikolov2013efficient} as follows:

 \begin{eqnarray} \label{eqn2}
 \mathcal{L}_{\mathcal{D}} &=& \sum_{(w, c) \in \mathcal{D}}\#(w, c)
 \Big(\log \sigma(\mathbf{w}\cdot \mathbf{c}) \nonumber
\\
 &+& \sum_{i=1}^k \mathbb{E}_{c'_i\sim P(w)}[\log \sigma(-\mathbf{w}\cdot \mathbf{c'}_i)]
 \Big)
 \end{eqnarray}
where $\mathcal{D}$ refers to the complete text corpus from which we learn the word embeddings. 
 The word $w$ is the current word, $c$ is the context word, and $\#(w,c)$ is the number of times they co-occur in $\mathcal{D}$.
 We use $\mathbf{w}$ and $\mathbf{c}$ to denote the vector representations for $w$ and $c$, respectively.
 The function $\sigma(\cdot)$ is the sigmoid function.
The word $c_i'$ is a ``negative sample'' sampled from the distribution $P(w)$ -- typically chosen as the unigram distribution $U(w)$ raised to the $3/4$rd power \cite{mikolov2013distributed}.

In our approach, we first learn for each word $w$ an embedding $\mathbf{w}_s$ from the source domain $\mathcal{D}_s$.
Next we learn the target domain embeddings as follows:
\begin{equation} \label{eqn1}
\mathcal{L}'_{\mathcal{D}_t} = \mathcal{L}_{\mathcal{D}_t} -  
\sum_{w\in\mathcal{D}_t\cap\mathcal{D}_s}
\alpha_w\cdot
||\mathbf{w}_t-\mathbf{w}_s||^2
\end{equation}
where $\mathcal{D}_t$ refers to the target domain, and $\mathbf{w}_t$ is the target domain representation for $w$.
Such an regularized objective can still be optimized using standard stochastic gradient descent.
Note that in the above formula, the regularization term only considers words that appear  in both source and target domain,
ignoring words that only appear in either the source or the target domain only.

Our approach is inspired by the recent regularization-based domain adaptation framework \cite{lugeneral}.
Here, $\alpha_w$ measures the amount of transfer across the two domains when learning the representation for word $w$. If it is large, it means we require the  embeddings of word $w$ in the two domains to be similar.
We define $\alpha_w$ as follows:

\begin{equation} \label{eqn4}
\alpha_w = \sigma(\lambda\cdot \phi(w)) 
\end{equation}
where $\lambda$ is a hyper-parameter to decide the scaling factor of the significance function $\phi(\cdot)$, which allows the user to control the degree of ``knowledge transfer'' from source domain to target domain. 

How do we define the significance function $\phi(w)$ that controls the amount of transfer for the word $w$?
We first define the frequency of the word $w$ in the dataset $\mathcal{D}$ as $f_{\mathcal{D}}(w)$, the number of times the word $w$ appears in the domain $\mathcal{D}$.
Based on this we can define the {\em normalized} frequency for the word $w$ as follows:

\begin{eqnarray}
\mathcal{F}_{\mathcal{D}}(w) = \frac{f_{\mathcal{D}}(w)}{\max_{w'\in \mathcal{D}_k} f_{\mathcal{D}}(w')} 
\end{eqnarray}
where $\mathcal{D}_k\subset\mathcal{D}$ consists of all except for the top $k$ most frequent words from $\mathcal{D}$\footnote{In all our experiments, we empirically set $k$ to 20.}.

We define the function $\phi(\cdot)$ based on the following metric that is motivated by the well-known S{\o}rensen-Dice coefficient \cite{sorensen1948method,dice1945measures} commonly used for measuring similarities:

\begin{eqnarray}
\phi(w) = \frac{2 \cdot \mathcal{F}_{\mathcal{D}_s}(w) \cdot \mathcal{F}_{\mathcal{D}_t}(w)}{\mathcal{F}_{\mathcal{D}_s}(w) + \mathcal{F}_{\mathcal{D}_t}(w)}
\end{eqnarray}

\begin{table*}[t!]
\small
\begin{center}
\begin{tabular}{l|cccccccc}
&Enwik9&PubMed&Gigaword (\textsc{En})&Yelp & IMDB&Tweets (\textsc{En})&Tweets (\textsc{Es})&Eswiki
\\
\hline
\# tokens&124.3M&124.9M&135.6M&38.9M&29.0M&162.8M&69.4M&102.8M\\

\# sents& -- &5,000,000&5,400,000&2,376,079&1,230,465&16,185,356&6,785,697&3,684,670\\
\end{tabular}
\end{center}
\label{default}
\vspace{-2mm}
\caption{Statistics for datasets used for embedding learning in all experiments.}
\label{tab:statistics}
\vspace{-4mm}
\end{table*}

Why does such a definition make sense? 
We note that the value of $\phi(w)$ would be high only if both both $\mathcal{F}_{\mathcal{D}_s}(w)$ and $\mathcal{F}_{\mathcal{D}_t}(w)$ are high -- in this case the word $w$ is a frequent word across different domains.
Intuitively, these are likely those words whose semantics do not change across the two domains, and we should be confident about making their embeddings similar in the two domains.
On the other hand, domain-specific words tend to be more frequent in one domain than the other. In this case, the resulting $\phi(w)$ will also have a lower score, indicating a smaller amount of transfer across the two domains.
While other user-defined significance functions are also possible, in this work we simply adopt such a function based on the above simple observations.
We will validate our assumptions with experiments in the next section.

\section{Experiments}
\label{exp}

We present extensive evaluations to assess the effectiveness of our approach.
Following recent advice by \citet{nayak2016evaluating} and \citet{faruqui2016problems}, to assess the quality of the learned word embeddings,
we considered employing the learned word embeddings as continuous features in several down-stream NLP tasks, including entity recognition, sentiment classification, and targeted sentiment analysis.

We have used various datasets from different domains for learning cross-domain word embeddings under different tasks.
We list the data statistics in Table \ref{tab:statistics}.

\subsection{Baseline Methods}


We consider the following baseline methods when assessing the effectiveness of our approach.

\begin{itemize}
\item \textsc{Discrete}: only discrete features (such as bag of words, POS tags, word $n$-grams and POS tag $n$-grams, depending on the actual down-stream task) were considered. All  following systems include both these base features and the respective additional features.
\item \textsc{Source}: we train word embeddings from the source domain as additional features.
\item \textsc{Target}: we train word embeddings from the target domain as additional features.
\item \textsc{All}: we combined the data from two domains to form a single dataset for learning word embeddings as additional features.
\item \textsc{Concat}: we simply concatenate the learned embeddings from both source and target domains as additional features.
\item DARep: we use the previous approach of \citet{bollegala2015unsupervised} for learning cross-lingual word representations as additional features.
\end{itemize}

\begin{figure*}[t!]
\centering
\scalebox{0.88}
{
\begin{tabular}{c}
\!\!\!\!\!
\begin{tikzpicture}
\begin{axis}[
    legend pos = north west,
    legend columns=2,
    legend style={font=\tiny, align=left,draw=none,column sep=0ex},
    xmin = 0.20, xmax = 1,
    ymin = 0.85, ymax = 0.92,
    xlabel = {Percentage of training set},
    ylabel = {$F_1$ score},
    height = {5cm},
    width = {8cm},
  enlargelimits=0.001, 
  grid style = {line width=0.5pt},
  ymajorgrids = true, 
  xmajorgrids = true, 
  ]
\addplot table [x index={0}, y index={1}, col sep=comma] {\imdbyelp}; 
\addplot table [x index={0}, y index={2}, col sep=comma] {\imdbyelp};
\addplot table [x index={0}, y index={3}, col sep=comma] {\imdbyelp};
\addplot table [x index={0}, y index={4}, col sep=comma] {\imdbyelp};
\addplot table [x index={0}, y index={5}, col sep=comma] {\imdbyelp};
\addplot table [x index={0}, y index={6}, col sep=comma] {\imdbyelp};
\addplot table [x index={0}, y index={7}, col sep=comma] {\imdbyelp};
\addplot table [x index={0}, y index={8}, col sep=comma] {\imdbyelp};

\legend{\textsc{Discrete},\textsc{Source},\textsc{Target},\textsc{All},\textsc{Concat},DARep,Lu et al.,This work};
\end{axis}
\end{tikzpicture}
\ \ \ \ 
\ \ \ \ 
\begin{tikzpicture}
\begin{axis}[
    legend pos = north west,
     legend columns=2,
    legend style={font=\tiny, align=left,draw=none,column sep=0ex},
    xmin = 0.20, xmax = 1,
    ymin = 0.76, ymax = 0.88,
    xlabel = {Percentage of training set},
    ylabel = {$F_1$ score},
    height = {5cm},
    width = {8cm},
    enlargelimits=0.001, 
    grid style = {line width=0.5pt},
    ymajorgrids = true, 
    xmajorgrids = true, 
    ]
\addplot table [x index={0}, y index={1}, col sep=comma] {\yelpimdb}; 
\addplot table [x index={0}, y index={2}, col sep=comma] {\yelpimdb};
\addplot table [x index={0}, y index={3}, col sep=comma] {\yelpimdb};
\addplot table [x index={0}, y index={4}, col sep=comma] {\yelpimdb};
\addplot table [x index={0}, y index={5}, col sep=comma] {\yelpimdb};
\addplot table [x index={0}, y index={6}, col sep=comma] {\yelpimdb};
\addplot table [x index={0}, y index={7}, col sep=comma] {\yelpimdb};
\addplot table [x index={0}, y index={8}, col sep=comma] {\yelpimdb};

\legend{\textsc{Discrete},\textsc{Source},\textsc{Target},\textsc{All},\textsc{Concat},DARep,Lu et al.,This work};
\end{axis}
\end{tikzpicture}
\end{tabular}
}
\captionsetup{justification=centering}
\vspace{-4mm}
\caption{Results on sentiment classification. Left: Yelp (source) to IMDB (target). Right: IMDB (source) to Yelp (target).}
\vspace{-4mm}
\label{imdbyelp}
\end{figure*}
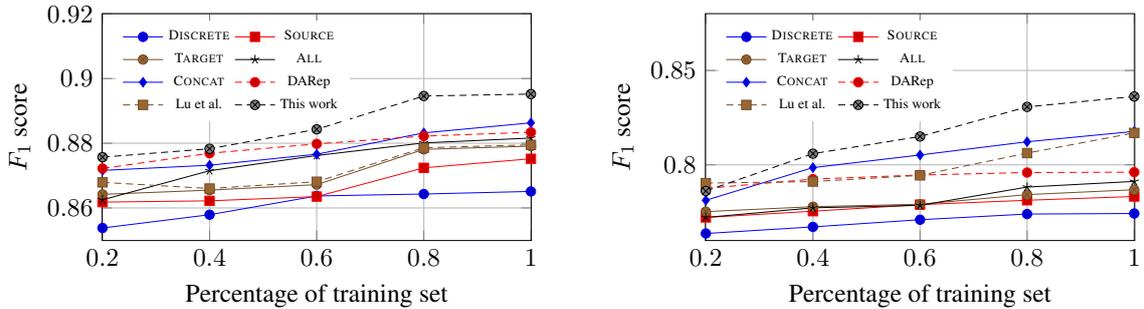

\subsection{Entity Recognition}

Our first experiment was conducted on entity recognition \cite{tjong2003introduction,florian2004statistical}, where the task is to extract semantically meaning entities and their mentions from the text.

For this task, we built a standard entity recognition model using conditional random fields \cite{lafferty2001conditional}.
We used the standard features which are commonly used for different methods, including word unigrams and bigrams, bag-of-words features, POS tag window features, POS tag unigrams and bigram features.
We conducted two sets of experiments on two different datasets.
The first dataset is the GENIA dataset \cite{ohta2002genia}, a popular dataset used in bioinformatics, and the second is the ACE-2005 dataset \cite{walker2006ace}, which is a standard dataset used for various information extraction tasks.

\begin{table}[t!]
\small
\centering
\begin{tabular}{l|ccc|ccc}
\multirow{2}{*}{Method}&\multicolumn{3}{c|}{GENIA}&\multicolumn{3}{c}{ACE}\\
&$P$&$R$&$F_1$&$P$&$R$&$F_1$
\\
\hline
\textsc{Discrete}&71.1&63.9&67.3&64.5&52.3&57.7\\
\hdashline
\textsc{Source}&71.1&62.3&66.4&63.5&57.3&60.3\\ 
\textsc{Target}&71.6&64.5&67.9&63.3&57.1&60.0\\ 
\textsc{All}&71.2&61.8&66.1&{\bf 64.6}&57.2&60.7\\ 
\textsc{Concat}&71.5&64.1&67.6&63.5&57.7&60.5\\ 
{DARep} &71.4&61.5&66.1&62.4&54.5&58.2\\ 
\hdashline
This work&{\bf 72.4}&{\bf 65.4}&\textbf{68.7}&64.5&\textbf{58.9}&\textbf{61.6}
\end{tabular}
\caption{Results on entity recognition.}
\vspace{-5mm}
\label{tab:results_NER}
\end{table}

For the GENIA dataset which consists of 10,946 sentences, we used Enwik9 as the source domain and PubMed as the target domain for learning word embeddings.
We set the dimension of word representations as 50.

For the experiments on ACE, we selected the BN subset of ACE2005, which consists of 4,460 CNN headline news and share a similar domain with Gigaword.
We used Enwik9 as the source domain and Gigaword as the target domain.
We followed a procedure similar to GENIA for experiments.

To tune our hyperparameter $\lambda$, we first split the last 10\% of the training set as the development portion.
We then trained a model using the remaining 90\% as the training portion and used the development portion for development of the hyperparameter $\lambda$.
After development, we re-trained the models using the original training set\footnote{We selected the optimal value for the hyper-parameter $\lambda$ from the set $\lambda\in\{0.1,1,5,10,20,30,50\}$ for all experiments in this paper.}.

We report the results in Table \ref{tab:results_NER}.
From the results we can observe that the embeddings learned using our algorithm can lead to improved performance when used in this particular down-stream NLP task.
We note that in such a task, many entities consist of domain-specific terms, therefore learning good representations for such words can be crucial.
As we have discussed earlier, our regularization method enables our model to differentiate domain-specific words from words which are more general in the learning process. 
We believe this mechanism can lead to improved learning of representations for both types of words.

\subsection{Sentiment Classification}


The second task we consider is sentiment classification, which is essentially a text classification task, where the goal is to assign each text document a class label indicating its sentiment polarity \cite{pang2002thumbs,liu2012sentiment}.

This is also the only task presented in the previous DARep work by \newcite{bollegala2015unsupervised}.
As such, we largely followed \newcite{bollegala2015unsupervised} for experiments.
Instead of using the dataset they used which only consists of 2,000 reviews, we considered two much larger datasets -- IMDB and Yelp 2014 -- for such a task, which was previously used in a sentiment classification task \cite{tang-qin-liu:2015:ACL-IJCNLP}.
{\color{black}IMDB dataset \cite{diao2014jointly} is crawled from the movie review site IMDB\footnote{{http://www.imdb.com}} {\color{black}which consists of 84,919 reviews. 
Yelp 2014 dataset consists of 231,163 online reviews provided by the Yelp Dataset Challenge\footnote{{https://www.yelp.com/dataset\_challenge}}}.}


Following  \newcite{bollegala2015unsupervised}, for this task we simply learned the word embeddings from the training portion of the review datasets themselves only.
No external data was used for learning word embeddings.
As \newcite{bollegala2015unsupervised} only evaluated on a small dataset in their paper for such a task, to understand the effect of varying the amount of training data, we also tried to train our model on datasets with different sizes.
We conducted two sets of experiments: we first used the Yelp dataset as the source domain and IMDB as the target domain, and then we switched these two datasets in our second set of experiments.
Figure \ref{imdbyelp} shows the $F_1$ measures for different word embeddings when different amounts of training data were used. 
We also compared with the previous approach for domain adaptation \cite{lugeneral} which only employs discrete features.
We can observe that when the dataset becomes large, our learned word embeddings are shown to be more effective than all other approaches.
When the complete training set is used, our model significantly outperforms DARep ($p<0.05$ for both directions with bootstrap resampling test \cite{koehn2004statistical}).
DARep appears to be effective when the training dataset is small. However, as the training set size increases, there is no significant improvement for such an approach.
As we can also observe from the figure, our approach consistently gives better results than baseline approaches (except for the second experiment when 20\% of the data was used).
Furthermore, when the amount of training data increases, the differences between our approach and other approaches generally become larger.

Such experiments show that our model works well when different amounts of data are available, and our approach appears to be more competitive when a large amount of data is available.

\begin{table}[t!]
\small
\centering
\begin{tabular}{l|ccc|ccc}
\multirow{2}{*}{Model} & \multicolumn{3}{c|}{English}  & \multicolumn{3}{c}{Spanish } \\
&   $P.$  & $R.$  & $F_1$ & $P.$  & $R.$  & $F_1$ \\
\hline
\textsc{Discrete} &44.8&37.0&40.5 &46.0&39.8&42.7\\
\hdashline
\textsc{Source} &44.1&36.3&39.8&46.1&40.5&43.1\\
\textsc{Target} &46.5&39.1&42.5&46.5&40.8&43.4\\
\textsc{All} &45.4&37.0&40.8&46.4&40.7&43.3\\
\textsc{Concat} &46.7&39.3&42.7&{\bf 46.6}&41.0&43.6\\
DARep&46.2&39.8&42.8&46.2&40.9&43.4\\
\hdashline
This work&\textbf{46.9}&\textbf{39.9}&\textbf{43.1}&\textbf{46.6}&\textbf{41.4}&\textbf{43.9} \\
\end{tabular}
\vspace{-3mm}
\caption{Results on targeted sentiment analysis.}
\vspace{-3mm}
\label{tab:TS}
\end{table}

\subsection{Targeted Sentiment Analysis}

We also conducted experiments on targeted sentiment analysis \cite{mitchell2013open} -- the task of jointly recognizing entities and their sentiment information.
We used the state-of-the-art system for targeted sentiment analysis by \newcite{lihao2017sentiment} whose code is publicly available \footnote{Available at http://statnlp.org/research/st/.}, and used the data from \cite{mitchell2013open} which consists of 7,105 Spanish tweets and 2,350 English tweets, with named entities and their sentiment information annotated.
Note that the model of \newcite{lihao2017sentiment} is a structured prediction model that involves latent variables.
The experiments here therefore allow us to assess the effectiveness of our approach on such a setup involving latent variables.
{\color{black}We follow \newcite{lihao2017sentiment} and report precision ($P.$), recall ($R.$) and F1-measure ($F_1$) for such a targeted sentiment analysis task, where the prediction is regarded as correct if and only if both the entity's boundary and its sentiment information are correct.}
Also, unlike previous experiments, which are conducted on  English only, these experiments additionally allow us to assess our approach's effectiveness when a different language other than English is considered.

For the English task, we used Enwik9 as the source domain for learning word embeddings, and our crawled English tweets as the target domain.
For the Spanish task, we used Eswiki as the source domain, and we also crawled Spanish tweets as the target domain.
See Table \ref{tab:statistics} for the statistics.
Similar to the experiments conducted for entity recognition, we split the first 80\% of the data for training, the next 10\% for development and the last 10\% for evaluation.
We tuned the hyper-parameter $\lambda$ using the development set and re-trained the embeddings on the dataset combining the training and the development set, which are then used in final evaluations.
Results are reported in Table \ref{tab:TS}, which show our approach is able to achieve the best results across two datasets in such a task, and outperforms DARep ($p<0.05$).
Interestingly, the concatenation approach appears to be competitive in this task, especially for the Spanish dataset, which appears to be better than the DARep approach.
However, we note such an approach does not capture any information transfer across different domains in the learning process. In contrast, our approach learns embeddings for the target domain by capturing useful cross-domain information and therefore can lead to improved modeling of embeddings that are shown more helpful for this specific down-stream task.

\section{Conclusion and Future Work}

In this paper, we presented a simple yet effective algorithm for learning cross-domain word embeddings.
Motivated by the recent regularization-based domain adaptation framework \cite{lugeneral}, the algorithm performs learning by augmenting the skip-gram objective with a simple regularization term.
Our work can be easily extended to multi-domain scenarios. 
The method is also flexible, allowing different user-defined metrics to be incorporated for defining the function controlling the amount of domain transfer.

Future work includes performing further investigations to better understand and to visualize what types of information has been transferred across domains and how such information influence different types of down-stream NLP tasks.
It is also important to understand how such an approach will work on other types of models such as neural networks based NLP models.
Our code is available at \mbox{\url{http://statnlp.org/research/lr/}}.

\section*{Acknowledgments}

We thank all the reviewers for their useful feedback to the earlier draft of this paper. 
This work was done when the first author was visiting Singapore University of Technology and Design.
We thank the support of Human-centered Cyber-physical Systems Programme at Advanced Digital Sciences Center from Singapores Agency for Science, Technology and Research (A*STAR).
This work is supported by MOE Tier 1 grant SUTDT12015008.

\bibliography{emnlp2017}
\bibliographystyle{emnlp_natbib}

\end{document}